\documentclass[journal]{IEEEtran}
\ifCLASSINFOpdf
\else
\fi
\usepackage{mathrsfs}
\usepackage{amsmath}
\usepackage{multirow}
\usepackage{graphicx}
\usepackage{color}
\usepackage{caption}
\usepackage{tabularx}
\usepackage{booktabs}
\usepackage{stfloats}
\usepackage{subfloat}
\usepackage{amsthm,amsmath,amssymb}
\usepackage{subfig}
\usepackage{amssymb}
\usepackage[citecolor=blue]{hyperref}

\begin{document}
\title{3DLG-Detector: 3D Object Detection via Simultaneous Local-Global Feature Learning}

\author{Baian~Chen, Liangliang Nan, Haoran Xie,~\IEEEmembership{Senior Member,~IEEE,}  Dening Lu, Fu Lee Wang,~\IEEEmembership{Senior Member,~IEEE,} and  Mingqiang Wei,~\IEEEmembership{Senior Member,~IEEE}
% <-this % stops a space
\thanks{B. Chen and M. Wei are with the School of Computer Science and Technology, Nanjing University of Aeronautics and Astronautics, Nanjing, China (e-mail: 2116068@nuaa.edu.cn;  mingqiang.wei@gmail.com).}
\thanks{L. Nan is with the Urban Data Science Section, Delft University of Technology, Delft, Netherlands (e-mail: liangliang.nan@tudelft.nl).}
\thanks{H. Xie is with the Department of Computing and Decision Sciences, Lingnan University, Hong Kong, China (e-mail: hrxie2@gmail.com).}
\thanks{D. Lu is with the Department of Systems Design Engineering, University of Waterloo, Waterloo, Canada (e-mail: d62lu@uwaterloo.ca).}

\thanks{F. L. Wang is with the School of Science and Technology, Hong Kong Metropolitan University, Hong Kong, China (e-mail: pwang@hkmu.edu.hk).}

}


\markboth{Journal of \LaTeX\ Class Files,~Vol.~14, No.~8, August~2015}%
{Shell \MakeLowercase{\textit{et al.}}: Bare Demo of IEEEtran.cls for IEEE Journals}



\maketitle

\begin{abstract}
Capturing both local and global features of irregular point clouds is essential to 3D object detection (3OD).
However, mainstream 3D detectors, e.g., VoteNet and its variants, either abandon considerable local features during pooling operations or ignore many global features in the whole scene context.
This paper explores new modules to simultaneously learn local-global features of scene point clouds that serve 3OD positively. To this end, we propose an effective 3OD network via simultaneous local-global feature learning (dubbed 3DLG-Detector).
3DLG-Detector has two key contributions. First, it develops a Dynamic Points Interaction (DPI) module that preserves effective local features during pooling. Besides, DPI is detachable and can be incorporated into existing 3OD networks to boost their performance. Second, it develops a Global Context Aggregation module to aggregate multi-scale features from different layers of the encoder to achieve scene context-awareness. Our method shows improvements over thirteen competitors in terms of detection accuracy and robustness on both the SUN RGB-D and ScanNet datasets. Source code will be available upon publication.
\end{abstract}


% \fi
\IEEEpeerreviewmaketitle

\section{Introduction}
\IEEEPARstart{R}{eal}-world complex scenes can be flexibly and efficiently represented by point clouds \cite{cad/YiLXLLWW19, wu2018automatic}.
3D object detection (3OD) in scene point clouds is a prerequisite for supporting the tasks like autonomous driving and augmented reality.
However, the captured point clouds of real-world scenes are natively irregular, compared to 2D (regular) images. Moreover, these point clouds are often sparse, incomplete, noisy, and contain outliers. Feature extraction from such irregularly-sampled yet degraded point clouds tends to weaken cutting-edging 3OD models to localize and recognize objects.

Representing the point clouds for effective processing in deep learning architectures is the first step for 3OD.
Currently, two representations have been widely used: voxel-based or point-based. 
The voxel-based methods \cite{8578570} divide a point cloud into regular 3D voxels and apply 3D CNNs to learn the high-level features. However, the memory and computational cost are growing exponentially with the increase in the resolution of voxels; it is also hard to trade-off between efficiency and accuracy for these methods \cite{8954311}. 

The point-based methods directly take raw point clouds as input to learn feature representations. To handle the irregularity of
point clouds without transformations, the seminal work of PointNet \cite{8099499} and PointNet++ \cite{NIPS2017_d8bf84be} apply multi-layer
perceptrons (MLPs) independently on each point, which enables to directly process sparse 3D points. 
Inspired by PointNet/PointNet++, 3D detectors \cite{8954080,9008567,9156597,9008777} have achieved satisfactory performance by designing various detection heads.

\begin{figure}[t]
    \centering
    \includegraphics[width=0.47\textwidth]{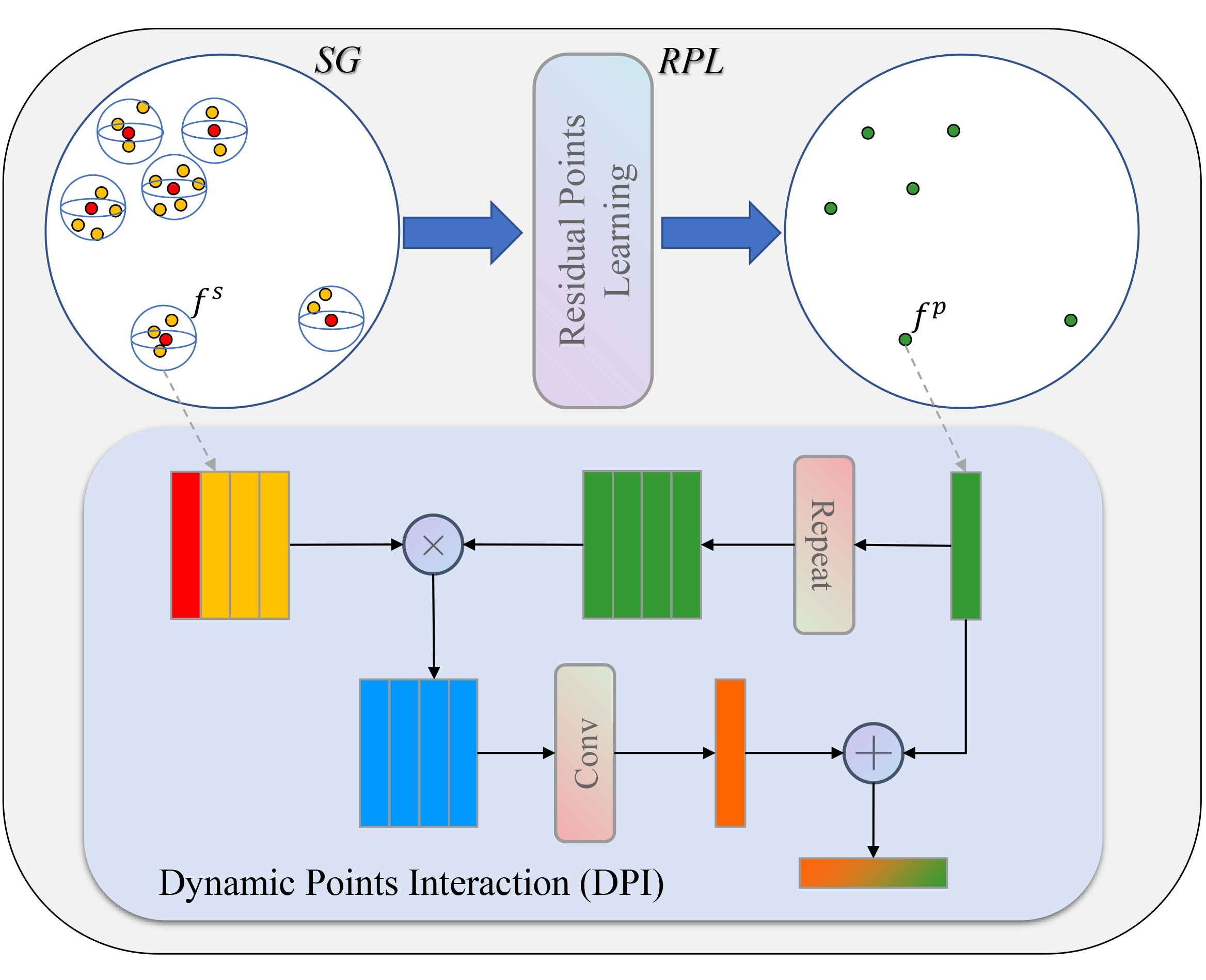}
    \caption{Dynamic Points Interaction, which can preserve the local features during pooling. Each feature encoder includes three sub-modules, i.e., the Sample-and-Group (SG) module, the Residual Points Learning (RPL) module, and the Dynamic Points Interaction (DPI) module. SG first performs a down-sampling operation on the input point cloud and then groups the neighbor points near the sampled point to form point set features $f^s$. These point sets are fed into RPL for deep feature representation learning. The max-pooling operation is used to aggregate the point set features into the seed as point-wise features $f^p$. DPI takes $f^s$ and $f^p$ as input, where $f^s$ is a supplement to make up the lost features of $f^p$ due to the max-pooling operation. }
    \label{fig:fig1}
\end{figure}

The key to 3OD is to simultaneously learn different scales and types of features from scene point clouds, such that the learned features can effectively capture both local geometric details and global scene features (context). 
The local features contribute to regressing the size and orientation of object bounding boxes, and the global features enhance inferring the classification of objects. 
The existing point-based 3D detectors learn point features based on PointNet/PointNet++, which inherit several drawbacks of PointNet/PointNet++.
First, utilizing PointNet/PointNet++ as backbones loses part of important local features. PointNet/PointNet++ utilizes simple symmetric functions, such as max-pooling, which is an indispensable component to deal with the permutation invariance for point cloud processing. However, the intrinsic character of max-pooling forces it to select the maximal value in each dimension as the representative feature. That means some equally important non-maximum features are lost in each dimension. \textit{We attempt to preserve such local features by designing a Dynamic Points Interaction module.}
Second, the global context can well describe the semantic information of the whole scene and the correlations between different objects in the scene. PointNet/PointNet++ only extracts the high-level feature representation by continuously expanding the receptive field while ignoring the global context.
The lack of global contextual information hinders the performance of these point-based detectors.
Although the recent Pointformer \cite{9578669} resorts to Transformer \cite{NIPS2017_3f5ee243} to learn the context-aware representation to capture the long-range dependency, it relies on plenty of data for long-term training, which is more difficult to train. \textit{We attempt to mine the global features by designing a Global Context Aggregation module.}

We propose 3DLG-Detector, a 3D object detection network by simultaneously learning local and global features. 
Inspired by dynamic learning \cite{NIPS2016_8bf1211f,10.1007/978-3-030-58452-8_17,9577670}, we design a Dynamic Points Interaction (DPI) module to preserve local features during pooling (see Figure \ref{fig:fig1}). In DPI, the input point cloud is first sampled and grouped to form a series of point sets. Then these point sets are fed to the Residual Points Learning module, which consists of several residual MLP blocks, to learn the deep feature representation and aggregate these point sets to seeds by the max-pooling operation.
The pooled seeds have simplified local context-aware features, while the grouped point sets possess detailed and redundant local geometric features. 
The DPI allows a seed to interact with each point in the corresponding point set to preserve local features.
Meanwhile, we observe that with the decreasing number of sampling points, the receptive field of each point in different encoder stages constantly increases. Hence, we design a Global Context
Aggregate (GCA) module to concatenate the multi-level features together to represent the contextual guidance. The final extracted features by GCA are therefore aware of the global information.

We conduct experiments on two indoor datasets, $i.e.$, ScanNet \cite{8099744} and SUN-RGBD \cite{7298655}. Extensive experiments have demonstrated the effectiveness of improvement under several evaluation metrics. 

In summary, our contributions are as follows:
\begin{itemize}
    \item We propose a novel 3D object detection network, 3DLG-Detector, which has a strong ability to learn local and global context features simultaneously. 
    Extensive experiments show clear improvements of our 3DLG-Detector over thirteen competitors in terms of both numerical and visual evaluations.
    \item We design three modules, among which the DPI and RPL modules extract rich local geometric information, and the GCA module captures the global scene context.
    Ablation experiments show the effectiveness of these modules in promoting detection performance. 
\end{itemize}

\section{Related Work}
\subsection{Feature Extraction for 3D Object Detection} 
\textbf{Local features extraction.}
Local feature extraction strategies can be divided into two categories: voxel-based and point-based methods. Voxel-based methods mostly use 3D sparse convolution \cite{8579059} on regular voxel grids. PointPillars \cite{8954311} directly adopts the mature 2D convolution by compressing the voxels into a pillar from the vertical dimension. For point-based methods, PointNet \cite{8099499} and PointNet++ \cite{NIPS2017_d8bf84be} directly consume unorganized 3D points and utilize symmetric functions and Set Abstract (SA) layers to learn the point-wise features and the local features progressively. PCCN \cite{8578372} exploits parameterized kernel functions to generalize convolution to learn the non-grid structured data. DGCNN \cite{10.1145/3326362} constructs a graph in the local region of sampled points and dynamically computes message propagation in each layer of the network. The above-mentioned strategies mostly use pooling operation for feature aggregation to progressively expand the receptive field of the sampled points, leading to the loss of local features. Similarly, our method extracts point-wise features from the local and global perspectives, respectively.
% \LL{add one sentence to briefly summarize the limitations of these methods and conclude that "These approaches unavoidably lose local features"}

\textbf{Multi-scale features learning.} With the continuously sampling operation on the point clouds, the perception field of each sampled point is extended incessantly. Multi-scale features are concatenated together as the overall scene information to ensure the local features are aware of the global context. PV-RCNN \cite{9157234} introduces the Voxel Set Abstraction (VSA) module to encode multi-scale voxel-wise features from the feature volumes to the key points. MLCVNet \cite{9156370} incorporates the multi-level context information from local point patches to global scenes into VoteNet. HVPR \cite{9578540} proposes an Attentive Multi-scale Feature Module (AMFM), which can refines the hybrid pseudo image to obtain scale-aware features.

\begin{figure*}[t]
    \centering
    \includegraphics[width=0.95\textwidth]{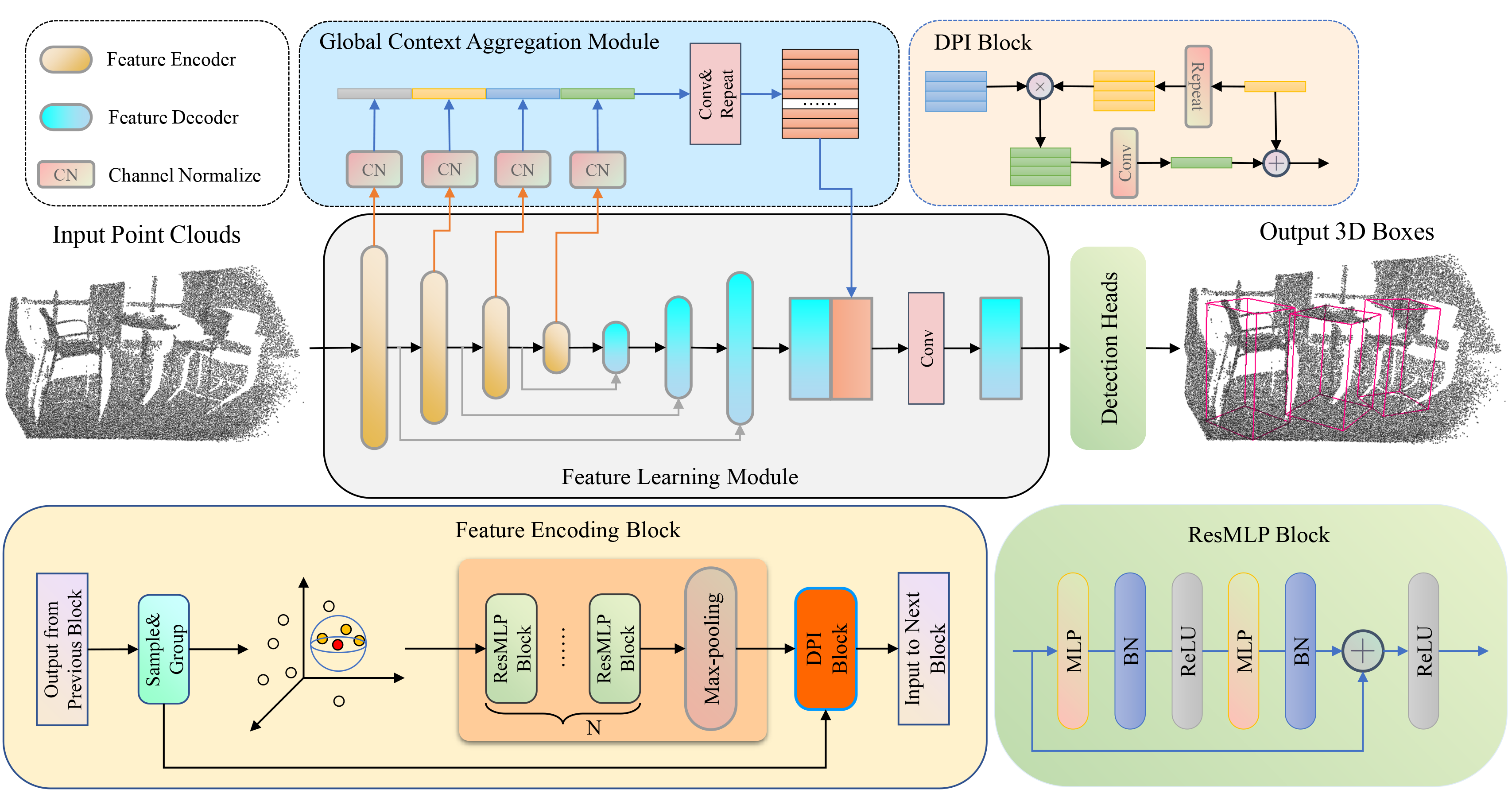}
    \caption{The pipeline of our 3DLG-Detector. The input scene point clouds are first fed into the feature learning module, in which a feature encoding (FE) block learns the high-level feature representations, and a feature decoding (FD) block recovers the discarded foreground points for accurate prediction. In each FE block, a Sample-and-Group (SG) module samples the seed points and groups the local region features near the seeds to expand the receptive field of the sampled points. Then, a Residual Points Learning (RPL) module further learns and aggregates the deep features. Finally, a Dynamic Points Interaction (DPI) module recovers the pooled local features. The outputs of FE at different levels have various receptive fields, which are concatenated together by a Global Context Aggregation (GCA) module as the global context to incorporate the global information into point features.}
    \label{fig:fig2}
\end{figure*}

\subsection{Voxel-based Object Detection} 
Voxel-based detectors first convert the point clouds into regular and compact voxel grids to utilize the matured convolutional neural networks. The current approaches can be divided into two groups: one-stage \cite{9156880,9196556,9157660,9157799,zheng2021cia} and two-stage detectors \cite{yin2021center,hu2022point,deng2021voxel,9018080}. The one-stage detectors focus on lightweight and efficiency, which usually lose the detailed structural information due to the voxelization and continuously down-sampling. SA-SSD \cite{9157660} introduces an auxiliary network to transform the convolution features into the point-wise representations to exploit the structural information. HVNet \cite{9157799} proposes a novel voxel feature encoder to attentively aggregate features at different levels and project the multi-scale feature maps to achieve accurate object localization. CIA-SSD \cite{zheng2021cia} proposes a lightweight aggregation module to fuse the semantic and spatial features to predict with accurate confidence, which is subsequently rectified by an IoU-aware rectification module. In comparison, the two-stage detectors pay more attention to the accuracy of detection. These detectors rely on the post-processing stage to refine the candidate proposals from the previous stage, which often has high demands in computation and memory. Voxel-RCNN \cite{deng2021voxel} exploits voxel RoI pooling to aggregate the voxel features within proposals for further refinement. Part-$A^2$ \cite{9018080} proposes a network with part-aware and part-aggregation stages, in which the former predicts proposals and locations of intra-object parts by the part supervision using ground truth boxes, and the latter excavates the spatial relationship of intra-object part locations to refine the proposals. 

\subsection{Point-based Object Detection} 
Point-based methods \cite{9710592,9711345,9710665,9008567} directly take point clouds, which keeps the original geometric information without any quantitative loss. However, it is challenging to achieve feature extraction due to the sparse and irregular characteristics of point clouds. PointRCNN \cite{8954080} is a two-stage 3D object detector, which first segments the foreground points and generates a small number of proposals. Then semantic features and local spatial cues are excavated from the proposals for further refinements.
% However, it is hard to further develop PointRCNN due to the expensive computation\LL{is efficiency a major concern, and is it addressed in this work? If not, I suggest removing this sentence}. 
VoteNet \cite{9008567} is a one-stage detector based on the Hough voting algorithm, which identifies instance centroids by voting from the points in a local region. Based on the VoteNet, MLCVNet \cite{9156370} proposes three context learning modules, respectively Patch-to-Patch Context, Object-to-Object Context, and Global Scene Context to capture the long-range dependencies at different levels. 3DSSD \cite{9156597} designs a novel fusion sampling strategy, which samples the farthest point according to the feature and Euclidean distance. Pointformer \cite{9578669} designs a transformer backbone to learn the context-dependent local features and context-aware global representations for 3D object detection.

The aforementioned point-based object detection methods mostly use PointNet++ as the backbone to extract features. However, their insufficient feature learning capacity limits the performance of the detectors. In this paper, we propose a novel feature learning framework for 3D object detection, which excavates and retains the complete local geometric cues by a dynamic points interaction module and captures the global scene context from different-level feature encoders.

 

\section{Methodology}

\subsection{Overview}
We propose a novel 3D object detector by learning both local and global features. It effectively preserves the local features after the pooling operation by dynamic points interaction and meanwhile learns the global context from multi-scale encoder blocks. As shown in Figure \ref{fig:fig2}, the feature learning module is an encoder-decoder structure. The feature encoder learns the high-level semantic features, in which the Sample-and-Group (SG) module first conducts points down-sampling and feature grouping. Then the Residual MLP (ResMLP) block learns a deeper feature representation from the grouped features. Lastly, the Dynamic Points Interaction (DPI) module takes the grouped features and pooled features as input and exploits grouped features to alleviate feature loss. The feature decoder follows the feature propagation module of PointNet++ to recover the discarded foreground points caused by downsampling. The outputs of the feature encoder blocks are concatenated together as global guidance, making point representations aware of the scene context. 

\begin{figure}[t]
    \centering
    \includegraphics[width=0.45\textwidth]{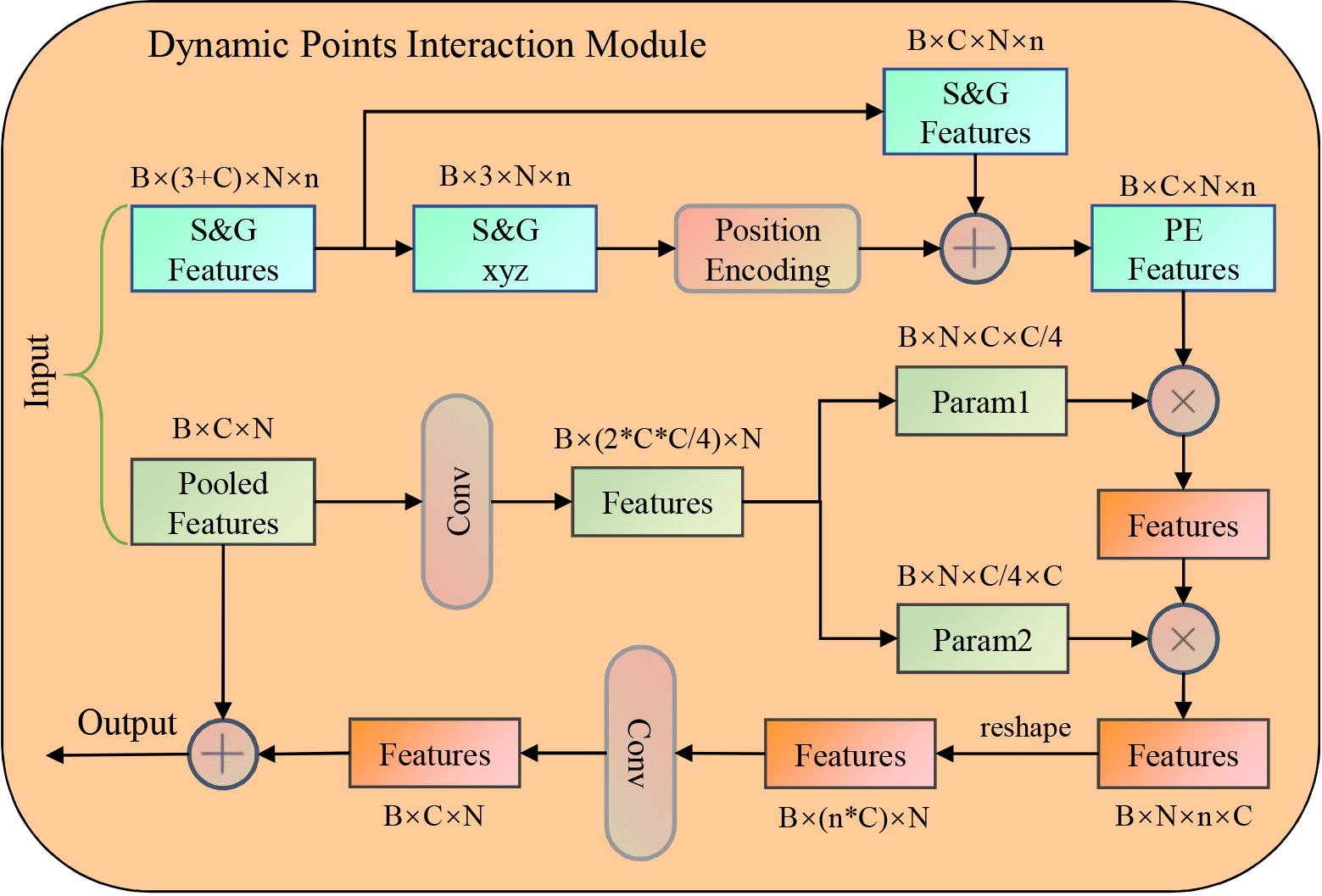}
    \caption{Illustration of the Dynamic Points Interaction module. The input includes grouped features and pooled features. The grouped features conduct position encoding to embed the position information as the query, and the pooled features are equally split to generate key-value pairs for carrying out dot-product with the query.}
    \label{fig:fig3}
\end{figure}

\subsection{Background}
The VoteNet \cite{9008567} is the baseline of our model, which consists of two components: the point features extraction module and the detection head. PointNet++ is the backbone network to extract high-level point features from the input point clouds. The detection head contains a voting module and a proposal module. The voting module takes the features from the previous component as input and regresses the offset from each seed point to the corresponding object center by MLPs, simulating the Hough voting process. The proposal module groups the predicted centers as object candidates to generate the 3D bounding boxes and classified labels.

In the succeeding works, MLCVNet \cite{9156370} reveals that the contextual information between different objects plays an active role in object recognization. Hence it designs three levels of context modules to learn the contextual information in the voting and proposal stages of VoteNet, respectively are Patch-to-Patch Context(PPC), Object-to-Object Context (OOC), and Global Scene Context (GSC) modules. Besides, Pointformer \cite{9578669} resorts to the popular Transformer to effectively learn context-aware feature representations. Specifically, a pointformer block consisting of the Local Transformer (LT) module and the Global Transformer (GT) module replaces the set abstract module of PointNet++ for feature extraction. However, these methods all neglect features loss during the pooling stage.

\subsection{Residual Points Learning Module}
The residual feed-forward MLPs have been proven to be effective for feature learning in PointMLP \cite{ma2021rethinking}. We introduce the Residual Points Learning (RPL) module by stacking the residual MLP blocks to learn the deeper point representations. As shown in Figure \ref{fig:fig2}, the RPL module can be formulated as
\begin{equation}
    g_i = \mathcal{A}(\phi{(f_{i,j})}|j=1,...,K),
    \label{eq:eq1}
\end{equation}
where $f_{i,j}$ is the feature of the $jth$ point near the $ith$ sampled point, and $\phi{(\cdot)}$ denotes the residual MLP block used to capture the deep feature. Specifically, the residual MLP block includes the mapping function $MLP(x)+x$, in which $MLP(\cdot)$ is combined by full connection, normalization, and activation layers. The aggregation function $\mathcal{A}$ is the max-pooling operation conducted on the features from the last residual MLP block to aggregate the local region features into the sampled point. Similar to ResNet \cite{he2016deep}, benefiting from the residual connections, the MLPs can easily be extended to dozens of layers for deeper feature representations. 

\subsection{Dynamic Points Interaction Module}
Although max-pooling operation leads to the loss of part of local geometric features, it is still an indispensable component in dealing with the permutation invariance for point cloud processing.
Thus we design a Dynamic Points Interaction (DPI) module to compensate for the feature loss caused by max-pooling without bypassing the max-pooling operation.
% In order to learn the complete local features and avoid the pooled features loss, we design a Dynamic Points Interaction (DPI) module to exploit the pooled seed to interact with each point in the corresponding grouped set for recovering the diminished part features. 
This dynamic interaction operation is similar to the attention function, which can be described as a mapping from the query term and key-value pairs to the output. The self-attention mechanism usually takes the same or similar features as input to perform QKV operations, focusing on excavating the inner relationship in features. In contrast, our dynamic points interaction (DPI) module takes grouped and pooled features as the query term and key-value pair. The QKV operation simulates the interaction process between pooled seeds and grouped sets and recovers the lost features progressively by continuous queries. Finally, the output is added with the pooled features to avoid pooled features being disturbed from the background point features in the grouped features.

The specific process is shown in Figure \ref{fig:fig3}, where the input of the DPI module includes the previous grouped features $F^g\in{R^{B\times{(3+C)\times{N\times{n}}}}}$ and pooled features $F^p\in{R^{B\times{C\times{N}}}}$. $F^g$ = \{$c_1,c_2,...,c_N$\}, where $c_i$ = \{$p_i,p_j,j=1,...,n-1$\} is a grouped points set in the local region. $p_i$ is a sampled point as the centroid of the set and $p_j$ is a neighboring point of $p_i$ within a given radius. Let $\{x_i,f_i\}_t$ denotes the point $p_i$ in the $t_{th}$ point set, where $x_i\in{R^3}$ represents the coordinates and $f_i\in{R^C}$ denotes the features of points.
Subsequently, the Position Encoding (PE) module takes $x_i$ as input to transform the dimension as the same of $f_i$ and adds $x_i$ to the $f_i$ in an element-wise manner for generating the queries $f_{q}$.
% coordinates are added with features in an element-wise manner after the Position Encoding (PE) operation to generate the PE features $f_{pe}$.\\
This process can be formulated as follows
\begin{equation}
    f_{q}=p_f\oplus{PE(p_{xyz})},
    \label{eq:eq2}
\end{equation}
where $p_{xyz}$ represents the coordinates and $p_f$ denotes the features of points.

The pooled features $F_p$ first carry out dimension extension from $C$ to $2*C*C/m$ by a convolution layer. Then these features are equally split into key-value pairs in the feature channel dimension, respectively are keys $f_k(:C*C/m)$ (Param1 in Figure \ref{fig:fig3}) and values $f_v(C*C/m:)$ (Param2 in Figure \ref{fig:fig3}). 
The reshape operation is adopted on the QKV features to change the arrangement of the feature dimension ($f_q\in{R^{B\times{N\times{n\times{C}}}}}$,$f_k\in{R^{B\times{N\times{C\times{(C/m)}}}}}$,$f_v\in{R^{B\times{N\times{(C/m)\times{C}}}}}$) to fit the succeeding Dot-Product function between queries and key-value pairs. To improve efficiency, we present a bottleneck structure between the key and value. We attempt to reduce the number of feature channels by a factor of $m$. In this paper, we set $m$ to 4. The whole calculation process can be formulated as follows,
\begin{equation}
    y = RB(RB(f_q\odot{f_k})\odot{f_v}),
    \label{eq:eq3}
\end{equation}
\begin{equation}
    o = R(y+F_p),
    \label{eq:eq4}
\end{equation}
where $W_q$, $W_k$, and $W_v$ are reshape operations for query, key, and value, respectively. $R$ and $B$ denote the activation function and the normalization function, respectively. 

The prior feature extraction modules in \cite{9578669,9156370} rely on the sophisticated feature extractor to excavate the local geometric information by using attention mechanisms. However, they do not design an effective strategy to preserve the extracted local features. The succeeding aggregation function (e.g., max-pooling) still inevitably deserts part important features.
% that proposes the Local Transformer (LT) to build hierarchical feature representations in the local region, our DPI module has obvious advantages.\LL{what advantages? Make them explicit} Even though the dense attention operation in LT has the capability of feature expressive and constructs the feature relationships in the local points set, the succeeding aggregation function (e.g., max-pooling) still inevitably deserts part important features. 
We give full consideration to this issue. The grouped set has redundant and comprehensive local features, in particular, including the part features lost by pooled seed.  
% The query and the key-value pairs in the DPI module come from the redundant and comprehensive grouped features and simplified pooled features respectively\LL{I cannot parse this sentence}. 
Hence we take the pooled seed continuously interacts with each point in the corresponding grouped set to acquire the completed local geometric information. 

\begin{table*}[t]
    \caption{3D object detection results on the ScanNet V2 validation set (left) and the SUN RGB-D validation set (right). The evaluation metric is the mean Average Precision with 3D IoU thresholds of 0.25 and 0.5. The results of the competing methods are quoted from their published papers or the released codes.}
    \begin{minipage}[t]{0.55\textwidth}
    \centering
    \makeatletter\def\@captype{table}
    \resizebox{\linewidth}{!}{
    \begin{tabular}{c|c|c|c}
        \toprule[1pt]
        ScanNet V2 & Input & mAP@0.25 & mAP@0.5 \\
        \midrule[0.8pt]
        DSS \cite{song2016deep} & Geo + RGB & 15.2 & 6.8    \\
        % MRCNN \cite{8237584} & Geo + RGB & 17.3 & 10.5  \\
        F-PointNet \cite{8578200} & Geo + RGB & 19.8 & 10.8  \\
        GSPN \cite{8953913} & Geo + RGB & 30.6 & 17.7 \\
        3D-SIS \cite{8954028} & Geo + 5 views & 40.2 & 22.5 \\
        \hline
        VoteNet \cite{9008567} & Geo only & 58.6 & 33.5 \\
        HGNet \cite{9156426} & Geo only & 61.3 & 34.4 \\
        DOPS \cite{9156417} & Geo only & 63.7 & 38.2 \\
        RGNet \cite{9234727} & Geo only & 48.5 & 26.0 \\
        MLCVNet \cite{9156370} & Geo only & 64.7 & 42.1 \\
        3DETR \cite{wang2022detr3d} & Geo only & 65.0 & 47.0 \\
        PointFormer \cite{9578669} & Geo only & 64.1 & 42.6 \\
        \hline
        Ours & Geo only & \textbf{66.3} & \textbf{48.0} \\
        \bottomrule[1pt]
    \end{tabular}
    }
    % \caption{Sample table title}
    % \label{sample-table}
    \end{minipage}
    \begin{minipage}[t]{0.45\textwidth}
    \centering
    \makeatletter\def\@captype{table}
    \resizebox{0.935\linewidth}{!}{
    \begin{tabular}{c|c|c}
        \toprule[1pt]
        SUN RGB-D & Input & mAP@0.25 \\
        \midrule[0.8pt]
        DSS \cite{song2016deep} & Geo + RGB & 42.1     \\
        2D-driven \cite{8237757} & Geo + RGB & 45.1      \\
        % PointFusion & Geo + RGB & 45.4 \\
        COG \cite{7780538} & Geo + RGB & 47.6 \\
        F-PointNet \cite{8578200} & Geo + RGB & 54.0 \\
        \hline
        VoteNet \cite{9008567} & Geo only & 57.7 \\
        H3DNet \cite{zhang2020h3dnet} & Geo only & 60.1 \\
        3DETR \cite{wang2022detr3d} & Geo only & 59.1 \\
        RGNet \cite{9234727} & Geo only & 59.2 \\
        MLCVNet \cite{9156370} & Geo only & 59.8 \\
        PointFormer \cite{9578669} & Geo only & 61.1 \\
        BRNet \cite{9578877} & Geo only & 61.1 \\
        \hline
        Ours & Geo only & \textbf{61.6} \\ 
        \bottomrule[1pt]
    \end{tabular}
    }
    % \caption{Sample table title}
    % \label{sample-table}
    \end{minipage}
    \label{tab:table1}
\end{table*}

\begin{table*}[t]\footnotesize
\renewcommand\arraystretch{1.2}
    \centering
    \caption{Results comparison with the state-of-the-art methods on the \textbf{ScanNetV2} validation set. The evaluation metric is the Average Precision with \textbf{0.5 IoU threshold.} 
    % `*' means the implementation of the model in MMDetection3D\LL{cite}.
    }
    \resizebox{\textwidth}{!}{
     \begin{tabular}{c|cccccccccccccccccc|c}
    \toprule[1pt]
        Methods & cab & bed & chair & sofa & table & door & wind & bkshf & pic & cntr & desk & curt & fridg & showr & toil & sink & bath & ofurn & mAP \\
    \midrule[0.8pt]
        VoteNet \cite{9008567} & 8.1 & 76.1 & 67.2 & 68.8 & 42.4 & 15.3 & 6.4 & 28.0 & 1.3 & 9.5 & 37.5 & 11.6 & 27.8 & 10.0 & 86.5 & 16.8 & 78.9 & 11.7 & 33.5 \\
    
        % VoteNet* & 14.6 & 77.9 & 73.1 & \textbf{80.5} & 46.5 & 25.1 & 16.0 & 41.8 & 2.5 & 22.3 & 33.3 & 25.0 & 31.0 & 17.6 & 87.8 & 23.0 & 81.6 & 18.7 & 39.9 \\
        
        % DOPS \cite{9156417} & \textbf{25.2} & 70.2 & 75.8 & 54.8 & 41.2 & 27.8 & 12.1 & 21.4 & 9.5 & 12.3 & 39.4 & 24.4 & 33.7 & 17.3 & 80.6 & 35.7 & 71.0 & 35.0 & 38.2 \\
         
        MLCVNet \cite{9156370} & 11.7 & 80.8 & 74.2 & 70.4 & 44.8 & 22.4 & 17.7 & 50.0 & 1.8 & 24.6 & 39.8 & 21.8 & 40.2 & 24.6 & 82.8 & 29.5 & 78.7 & 17.2 & 40.7 \\
        
        Pointformer \cite{9578669} & 19.0 & 80.0 & 75.3 & 69.0 & 50.5 & 24.3 & 15.0 & \textbf{41.9} & 1.5 & 26.9 & 45.1 & 30.3 & 41.9 & 25.3 &  75.9 & 35.5 & 82.9 & 26.0 & 42.6 \\
    
        3DETR \cite{wang2022detr3d} & \textbf{24.4} & 79.4 & 76.5 & 67.8 & 53.0 & 25.7 & 15.7 & 41.8 & 6.1 & 20.8 & \textbf{46.8} & 26.7 & 37.8 & \textbf{40.1} & \textbf{96.0} & 30.2 & 84.4 & 28.3 & 44.5 \\
    
    \hline
        Ours & 20.0 & \textbf{80.6} & \textbf{79.1} & \textbf{77.7} & \textbf{61.3} & \textbf{34.1} & \textbf{21.7} & 41.2 & \textbf{10.8} & \textbf{28.3} & 39.0 & \textbf{34.8} & \textbf{54.3} & 34.6 & 90.4 & \textbf{36.1} & \textbf{88.3} & \textbf{31.9} & \textbf{48.0} \\
    \bottomrule[1pt]
    \end{tabular}
    }
    \label{tab:table2}
\end{table*}
%\IEEEpubidadjcol

\subsection{Fourier Position Encoding}
Position encoding is an essential component of Transformer since it can embed relative or absolute position information of each entry in the input to the corresponding features. For 3D point clouds, the position information of each point still plays a crucial role in describing the local geometric structure of the point clouds. 

Inspired by \cite{tancik2020fourier,he2022voxel}, we introduce a Fourier Position Encoding to map the low-dimension coordinates to the higher frequency representations by the heuristic sinusoidal function. Specifically, the function $\gamma$ maps the coordinates ($xyz\in[0,1]$) of the input points to the higher dimensional hypersphere with a set of sine-cosine functions
\begin{equation}
    \delta_i{(v)} = (a_icos(2\pi{b_iv}),a_isin(2\pi{b_iv})),
\end{equation}
\begin{equation}
    \gamma{(v)} = [\delta_1{(v)},...,\delta_m{(v)}], v\in\{x,y,z\},
\end{equation}
where $b_i$ is the Fourier basis frequency and $a_i$ is the corresponding Fourier series coefficient. For simplicity, we set $a_i$ = 1 and generate $b_i$ by a power function $b_i$ = $T^{i/m}$, $i=0,...,m-1$. The results from the Fourier embedding are concatenated together as position encoding with a dimension of $3m$, and they are further transformed such that their dimension is the same as the corresponding point features.

\subsection{Global Context Aggregation Module}
The global context describes the semantic information of the whole scene, which is considerable in inferring the classes of objects as there is a close connection between the scene and objects. Prior works, no matter point-based models using PointNet++ or voxel-based methods using sparse 3D convolution, only extract the high-level feature representation by continuously expanding the receptive field but neglect the global context.

We note that the high-level features include rich semantic information while the low-level features contain the local geometric cues. Hence we propose the global context aggregation (GCA) module to concatenate them together as the global context guidance, to promote the ability of feature representations for 3D bounding box regression and object classification. Specifically, we first conduct the channel normalization (CN) to the outputs of each feature encoding block. This operation is to compress the number of the feature channel to $k$ for the succeeding concatenation. The formulation of CN can be summarized as follows:
\begin{equation}
    CN(f) = Max-Pooling(MLP(f)),
    \label{eq:eq5}
\end{equation}

To solve the problem of the inconsistent number of the sampled points from different encoders, the max-pooling function is applied to compress the features to a 1D vector. Subsequently, these vectors representing respective encoders are concatenated together as the global context,
\begin{equation}
    g = MLP(Cat[CN(f_i), i=1,2,3,4]).
\end{equation}

The global context representations not only promote the message propagation among different objects in the scene, but also benefit the inference in object classification. %In this way, the final detection results will be more reliable and effective.

\section{Experiments}
In this section, we conduct extensive experiments on two indoor datasets to evaluate the proposed 3DLG-Detector and compare it with the state-of-the-art 3D object detection methods. In Section \ref{sec:4.1}, we introduce the details of datasets and the setup of the model. In section \ref{sec:4.2}, we demonstrate the qualitative and quantitative comparison results on indoor datasets. In section \ref{sec:4.3}, we analyze the effectiveness of each component in 3DLG-Detector through comprehensive ablation studies. In section \ref{sec:4.4}, we introduce the limitation of our model by analyzing several failure cases.

\subsection{Datasets and Implementation Details}
\label{sec:4.1}
We evaluate our method on two indoor datasets, SUN RGB-D \cite{7298655} and ScanNet V2 \cite{8099744}. 

\textbf{SUN RGB-D} \cite{7298655} is a single-view RGB-D dataset for 3D scene understanding. It contains $\sim$ 5K indoor RGB and depth images annotated with amodal oriented bounding boxes of 37 object categories for training, and the rest $\sim$ 5K RGB-D images for testing. Before feeding the data into the network, depth images are first converted to point clouds by the provided camera parameters. The evaluation metric is the standard mean Average Precision (mAP), and the evaluation is conducted on the 10 most common categories. 

\textbf{ScanNet V2} \cite{8099744} is a densely annotated dataset consisting of 3D reconstructed meshes, which has rich texture, semantic and geometric information. It contains 1513 indoor scenes captured from hundreds of different rooms, with semantic and instance labels for all the points, as well as 3D object bounding boxes. Compared to the fragmentary scan in SUN RGB-D, the scenes of ScanNet are larger and more complete, so local geometric details of objects are well captured. The vertices of the meshes in the dataset are sampled as point clouds.

\textbf{Data augmentation.} To reduce computational complexity, we randomly down-sample each point cloud as input, $i.e.$, 20,000 points for the SUN RGB-D dataset and 40,000 points for the ScanNet dataset respectively. The height attribute of each point is also included as an extra feature to feed into the network. To augment the training data, we apply randomly flipping, rotating, and scaling operations to the point clouds, following VoteNet \cite{9008567}. 

\textbf{Training details.} Our model is implemented with PyTorch on an NVIDIA GeForce RTX 3060 GPU and optimized by the Adam optimizer in an end-to-end manner. For ScanNet V2, we set the initial learning rate to 1e-3 and weight decay to 1e-1. The total training epochs are 48, and the learning rate continuously decreases in the 12, 24, and 36 epochs by $5\times$. For the other dataset SUN RGB-D, we set the base learning rate to 1e-3 and weight decay to 5e-2. The total epochs are 36, and the learning rate continuously decreases in the 12 and 24 epochs by $5\times$.

\begin{figure*}[t]
    \centering
    \includegraphics[width=0.95\textwidth]{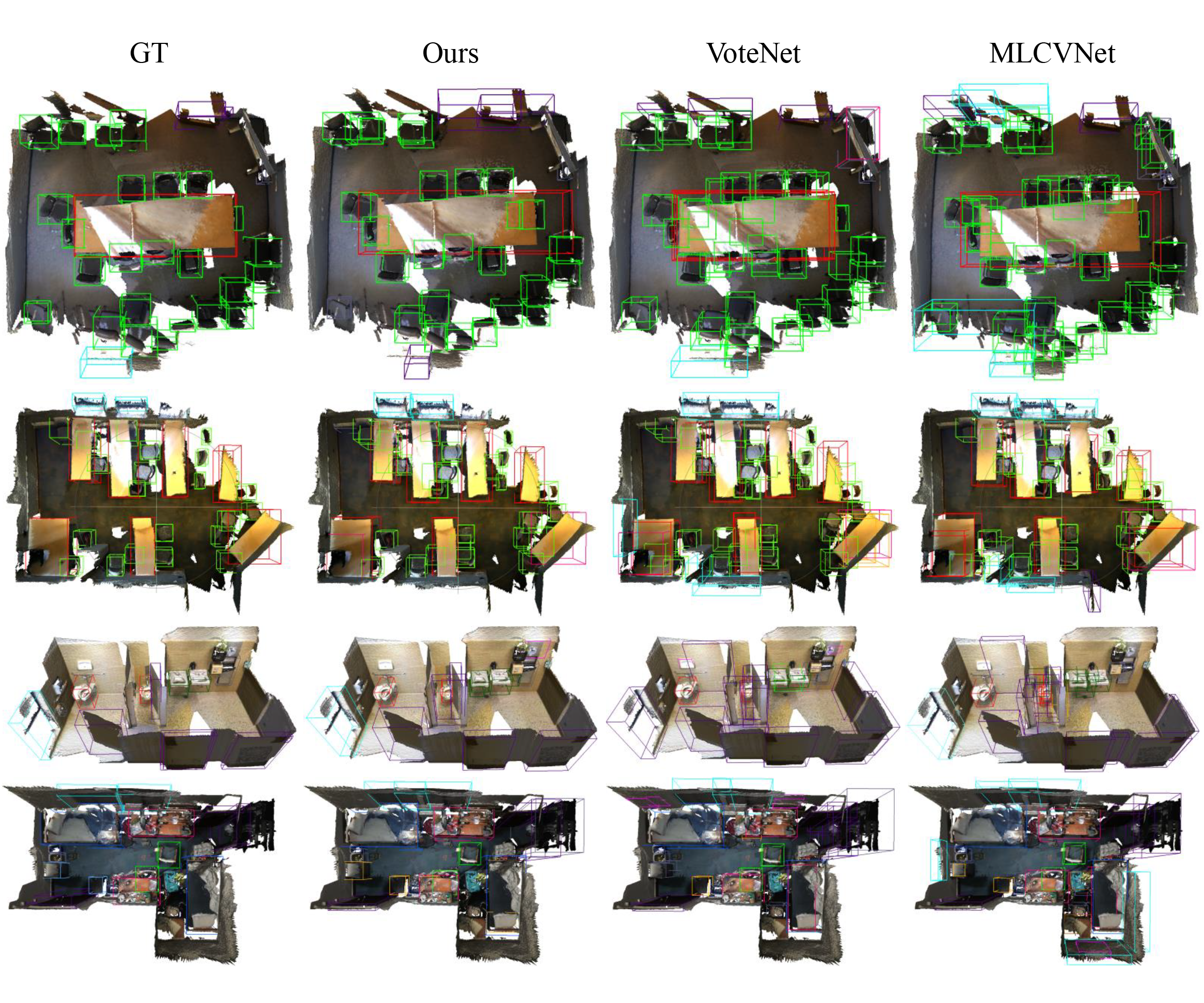}
    \caption{The qualitative results of different 3D object detection methods on the ScanNet V2 validation sets. }
    \label{fig:fig4}
\end{figure*}

\subsection{Comparisons with the State-of-the-art Methods}
\label{sec:4.2}
We compare our method with the related works, which can be divided into three groups: early methods \cite{8954028,8237757,8578200,8953913,7780538} that locate 3D objects via 3D-2D queries, voting-based methods that excavate informative local representation such as VoteNet \cite{9008567} and its successors \cite{9156426,9156417,9578877,zhang2020h3dnet}, and attention-based methods \cite{9234727,wang2022detr3d,9578669,9156370} that explore the relationships between the local objects and point clusters.
The results are reported in Table \ref{tab:table1} and Table \ref{tab:table2}. The bold texts denote the best results under the corresponding metrics. 

\begin{figure*}[t]
    \centering
    \includegraphics[width=0.95\textwidth]{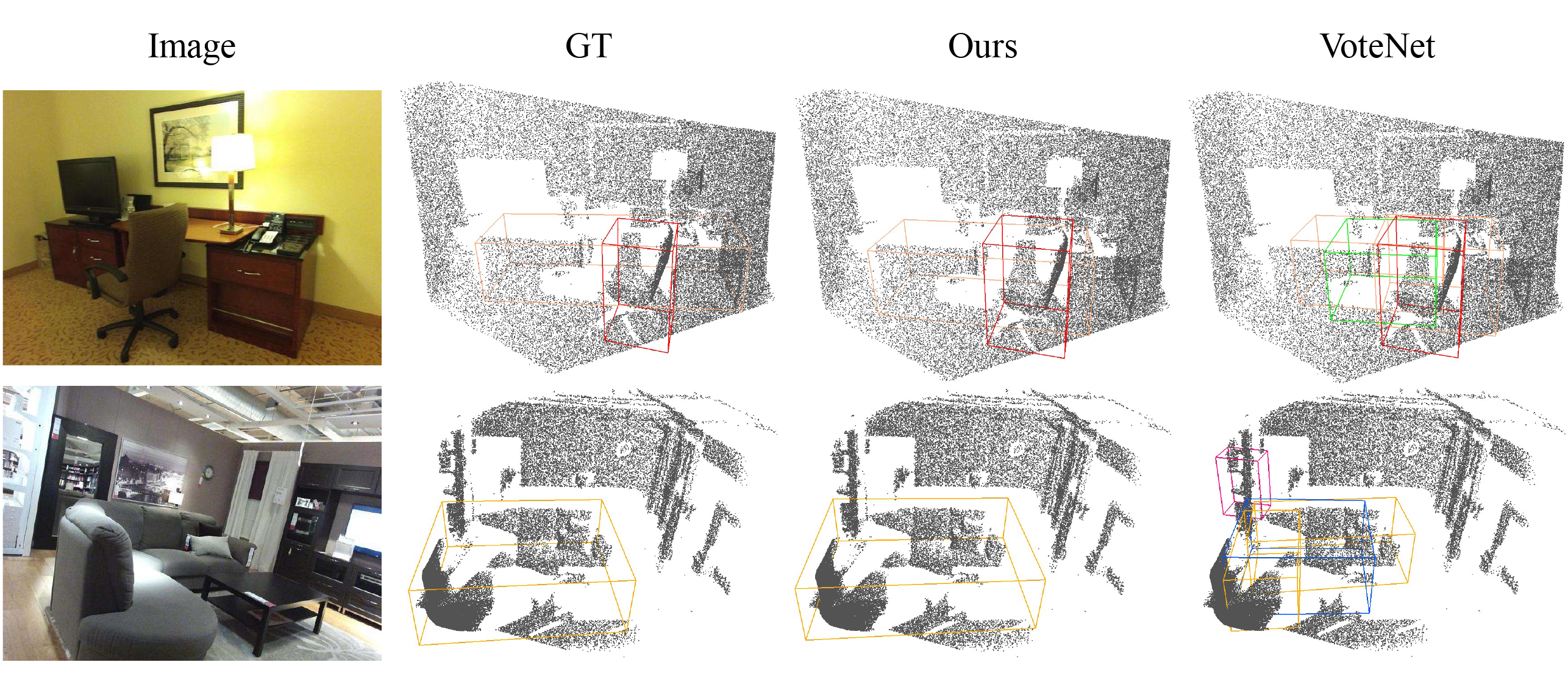}
    \caption{The qualitative results of different 3D object detection methods on the SUN RGB-D validation sets. }
    \label{fig:fig5}
\end{figure*}

\textbf{Quantitative results.} All results of comparison experiments are summarized in Table \ref{tab:table1}. We can observe that 3DETR \cite{wang2022detr3d} has the highest mAP among the competing methods on the ScanNet dataset, while our method still outperforms 3DETR in both metrics on the ScanNet V2 validation sets (+1.3\% $mAP@0.25$, +1\% $mAP@0.5$). Note that $mAP@0.5$ is quite a challenging metric since it requires more than 79\% coverage area in each dimension of the bounding box. Some methods like HGNet \cite{9156426} only achieve decent performance under the metric $mAP@0.25$ but perform exceptionally poorly in terms of the $mAP@0.5$ metric. Our model has the highest accuracy in object location, so it retains significant performance. 
The ScanNet dataset consists of reconstructed meshes that cover complete objects in larger areas, while the SUN RGB-D dataset contains the single-view RGB-D images where point clouds projected from the depth map have fragmentary objects and smaller areas.
% \LL{I cannot understand this sentence: what is "projected point cloud" and what does "small" means (in terms of area or number of points). Rephrase it to be specific.}
The different characteristics result in the inability of many methods to perform consistently well on both datasets. For example, MLCVNet \cite{9156370} performs well on the ScanNet dataset but achieves poor results on the SUN RGB-D dataset, while RGNet \cite{9234727} is the opposite. Our method has also achieved impressive performance on the SUN RGB-D dataset, indicating it has a strong generalization capability to deal with different scenes.

The comparison results on the ScanNet dataset in terms of $mAP@0.5$ are shown in Table \ref{tab:table2}. Our method achieves the best performance in 13 out of the 18 categories. Especially for the tabular objects like pictures and windows, whose neighborhoods mostly are background points. Other methods cannot detect them due to that the pooling operation aggregates too many background features but discards important object features. Nevertheless, our dynamic points interaction module preserves object features, which improves the detection accuracy by 4\% AP and 4.1\% on windows and pictures.

\textbf{Qualitative results.} We visualize the representative detection results from the ScanNet dataset and SUN RGB-D dataset in Figure \ref{fig:fig4} and Figure \ref{fig:fig5}, from which we can observe that the VoteNet \cite{9008567} and MLCVNet \cite{9156370} have wrong detection regarding object number and category. For example, in the first row of Figure \ref{fig:fig4}, VoteNet \cite{9008567} and MLCVNet \cite{9156370} recognize many wrong chairs on the table and in the wall. In contrast, enhanced by the proposed DPI and GCA modules, our model achieves more accurate bounding boxes in terms of both location and category.

\begin{table}[]
\renewcommand\arraystretch{1.2}
    \centering
    \caption{Ablation experiments regarding the number of ResMLP blocks in the RPL module. Note `0*' denotes using the traditional MLPs.}
    \begin{tabular}{c|c|c}
    \toprule[1pt]
    \multirow{2}{*}{ResMLP} & \multicolumn{2}{c}{ScanNet V2} \cr\cline{2-3}
    & mAP@0.25 &mAP@0.5 \cr
    \hline
        0* block & 63.9 & 45.4 \\
        1 block & 64.9 & 47.1 \\
        2 blocks & \textbf{66.3} & \textbf{48.0} \\
        3 blocks & 65.2 & 47.0 \\
    \bottomrule[1pt]
    \end{tabular}
    \label{tab:table3}
\end{table}

\begin{table}[t]
\renewcommand\arraystretch{1.2}
    \centering
    \caption{Ablation experiments regarding the number of ResMLP blocks in the RPL module. `-DPI' means the 3DLG-Detector without the DPI module, `-GCA' indicates the 3DLG-Detector without the GCA module.}
    \resizebox{\linewidth}{!}{
    \begin{tabular}{c|c|c|c|c}
    \toprule[1pt]
    \multirow{2}{*}{ResMLP} & \multicolumn{2}{c|}{ScanNet V2} & \multicolumn{2}{c}{SUN RGB-D} \cr\cline{2-5}
    & mAP@0.25 & mAP@0.5 & mAP@0.25 & mAP@0.5 \cr
    \hline
        VoteNet & 58.6 & 33.5 & 57.7 & 32.9 \\
        -DPI & 64.7 & 45.4 & 58.5 & 35.1 \\
        -GCA & 65.3 & 46.1 & 60.1 & 37.6 \\
        3DLG-Detector & \textbf{66.3} & \textbf{48.0} & \textbf{61.6} & \textbf{38.5} \\
    \bottomrule[1pt]
    \end{tabular}
    }
    \label{tab:table4}
\end{table}

\begin{table}[t]
    \centering
    \caption{Ablation experiments about comparing DPI module with self-attention.}
    \resizebox{\linewidth}{!}{
    \begin{tabular}{c|c|c|c|c}
    \toprule[1pt]
    \multirow{2}{*}{ResMLP} & \multicolumn{2}{c|}{ScanNet V2} & \multicolumn{2}{c}{SUN RGB-D} \cr\cline{2-5}
    & mAP@0.25 & mAP@0.5 & mAP@0.25 & mAP@0.5 \cr
    \hline
        self-attention & 64.7 & 45.0 & 60.1 & 36.3 \\
        DPI & \textbf{66.3} & \textbf{48.0} & \textbf{61.6} & \textbf{38.5} \\
    \bottomrule[1pt]
    \end{tabular}
    }
    \label{tab:table5}
\end{table}

\begin{figure}[t]
    \centering
    \includegraphics[width=0.45\textwidth]{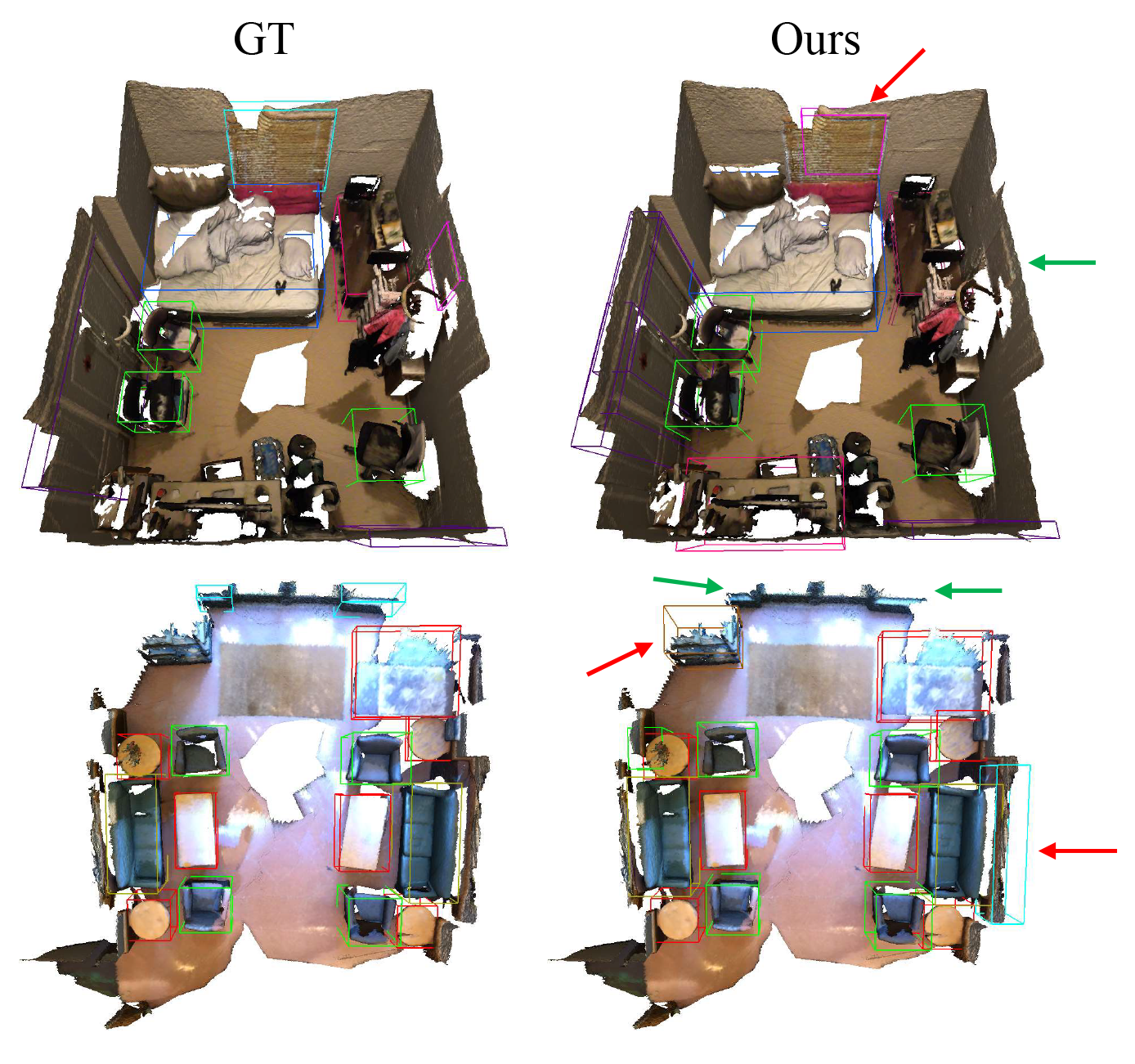}
    \caption{Examples of failure cases on the ScanNet V2 dataset. The red arrows denote the false positive bounding boxes, and the green arrows indicate the missed objects.}
    \label{fig:fig7}
\end{figure}

\subsection{Ablation Study}
\label{sec:4.3}
\textbf{Residual points learning module.} We first evaluate the effect of the number of ResMLP blocks in the Residual Points Learning (RPL) module on feature learning. We change the depth of the RPL module by setting the number of ResMLP blocks to 0, 1, 2, and 3, respectively. 0 block means using the traditional MLP layer for feature extraction. The experiment results are reported in Table \ref{tab:table3}, from which we observe an increase in detection performance as the RPL module becomes deeper. However, merely increasing the number of ResMLP blocks would not always lead to better performance. When setting the number of ResMLP blocks to 3, the detection accuracy decreases $mAP@0.25$ by 1.1\% and $mAP@0.5$ by 1.0\%. In this work, two ResMLP blocks achieve the best performance.  

\begin{figure*}[t]
    \centering
    \includegraphics[width=0.95\textwidth]{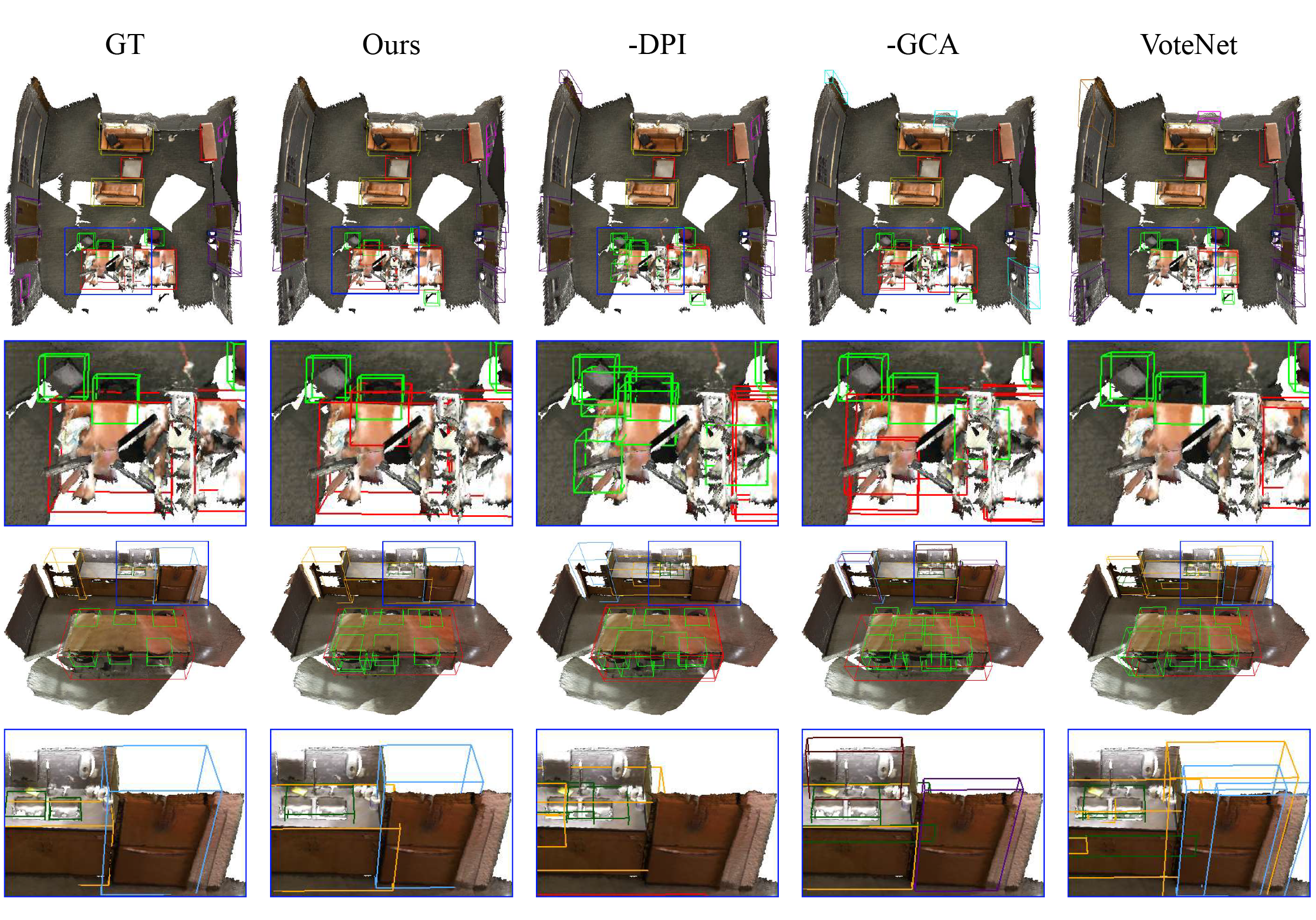}
    \caption{The visual results of ablation experiments on the ScanNet V2 validation sets. `-DPI' denotes the 3DLG-Detector without the DPI module, `-GCA' indicates the 3DLG-Detector without the GCA module. The first and third rows demonstrate the whole scenes, and the second and fourth rows are close-up views.}
    \label{fig:fig6}
\end{figure*}

\textbf{Dynamic points interaction module.} DPI module is the essential component in our model, which significantly improves detection accuracy. The quantitative results are reported in Table \ref{tab:table4}. We can see that without the DPI module, the performance drops 2.6\% and 3.4\% in terms of $mAP@0.5$ on the ScanNet and SUN RGB-D validation sets, respectively. The visualization of the object detection results is in the first two rows of Figure \ref{fig:fig6}. After removing the DPI module, several chairs (green boxes) instead of the table (red boxes) are incorrectly detected. This is also due to that the pooling operation aggregates features from the neighbor regions instead of object features. The sampled points of the table integrate with the points from the chairs beside the table, leading to the error of recognizing the table as several chairs. Our DPI module enables the grouped features to interact with the pooled features to preserve local features and thus ensures correct table detection.

\textbf{Differences between DPI and self-attention.} Although our dynamic points interaction module is similar to the self-attention in formulation, the inputs are completed different. Self-attention takes same or similar features as input while DPI takes grouped and pooled features as the query term and key-value pair. We apply the self-attention on the pooled features to replace the DPI module. We can find the performance has decreased by 3.0\% and 2.2\% in terms of $mAP@0.5$ on the ScanNet and SUN RGB-D validation sets, respectively. The reason is that self-attention only excavates the internal relationship of features while neglecting to introduce the external cues to compensate for the feature loss. 

\textbf{Global context aggregation module.} GCA module plays a substantial role in learning the global contextual information for 3D object detection. As shown in Table \ref{tab:table4}, removing the GCA module causes the detection accuracy to decrease by 1.9\% and 0.9\% in terms of $mAP@0.5$ on the ScanNet and SUN RGB-D validation sets, respectively. The visualization results are shown in the last two rows of Figure \ref{fig:fig6}. The fridge near the sink is wrongly detected as a door by the model without the GCA module. The global scene context encodes the multi-scale features to generate scene context information that helps to enhance object detection. 

\subsection{Limitations}
\label{sec:4.4}
Although 3DLG-Detector has demonstrated notable improvement on two indoor datasets, it still does not perform well for a few tricky scenes. Two such failure cases are presented in Figure \ref{fig:fig7}. The common failures are false-positive bounding boxes of objects (red arrows in Figure \ref{fig:fig7}) and the missed object detection (green arrows in Figure \ref{fig:fig7}). We can see that the picture and the window in the smooth wall are the most challenging to be detected because they are too thin and clung to the wall. The false-positive bounding boxes also arise when several objects have similar shapes, for which the global context cannot distinguish between them. It is worth noting that these common failures are equally problematic for most SOTA methods. Additionally, we have not resolved the over smoothing phenomenon caused by the residual points learning module. The residual connection can deepen the MLPs from several to dozens of layers to learn better deep feature representations. However, the performance may not be enhanced and may degrade when the number of layers increases largely.

\section{Conclusion}
We have presented a novel framework to improve voting-based 3D object detection networks. Our approach enhances the learning of both local and global features by introducing three different modules to the networks. The RPL module first learns the deep local feature representation, and then the DPI module captures the complete local geometric features. The GCA module constructs global contextual information from multi-scale feature encoders, thus enriching global features. Extensive experiments have demonstrated the effectiveness of the proposed approach. 

Compared to prior works that propose sophisticated feature extractors to excavate detailed local geometric information, our work takes a different path to preserve the extracted features. The flourishing extractors have saturated performance in describing local geometric information while designing effective feature retention strategies has been rarely studied. We believe our work can promote the research in feature retention.

% or


% \appendices

% % you can choose not to have a title for an appendix
% % if you want by leaving the argument blank
% \section{}

% % use section* for acknowledgment
% \section*{Acknowledgment}


\ifCLASSOPTIONcaptionsoff
  \newpage
\fi



%\bibliographystyle{IEEEtran}
\bibliographystyle{IEEEtran} 
\bibliography{ref} 



% \end{thebibliography}


\begin{IEEEbiography}[{\includegraphics[width=1in,height=1.25in,clip,keepaspectratio]{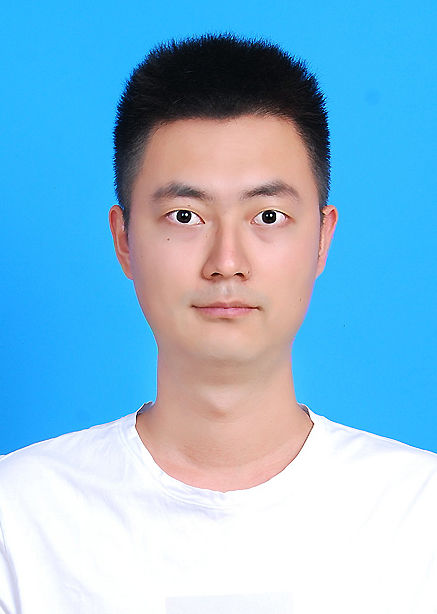}}]{Baian Chen}
 	is now pursuing his PhD degree at Nanjing University of Aeronautics and Astronautics (NUAA), China. He received his B.Sc. degree from China University of Mining and Technology. His research interests include 3D vision and learning-based geometry processing.
 \end{IEEEbiography}

\vspace{-10 mm}
\begin{IEEEbiography}[{\includegraphics[width=1in,height=1.25in,clip,keepaspectratio]{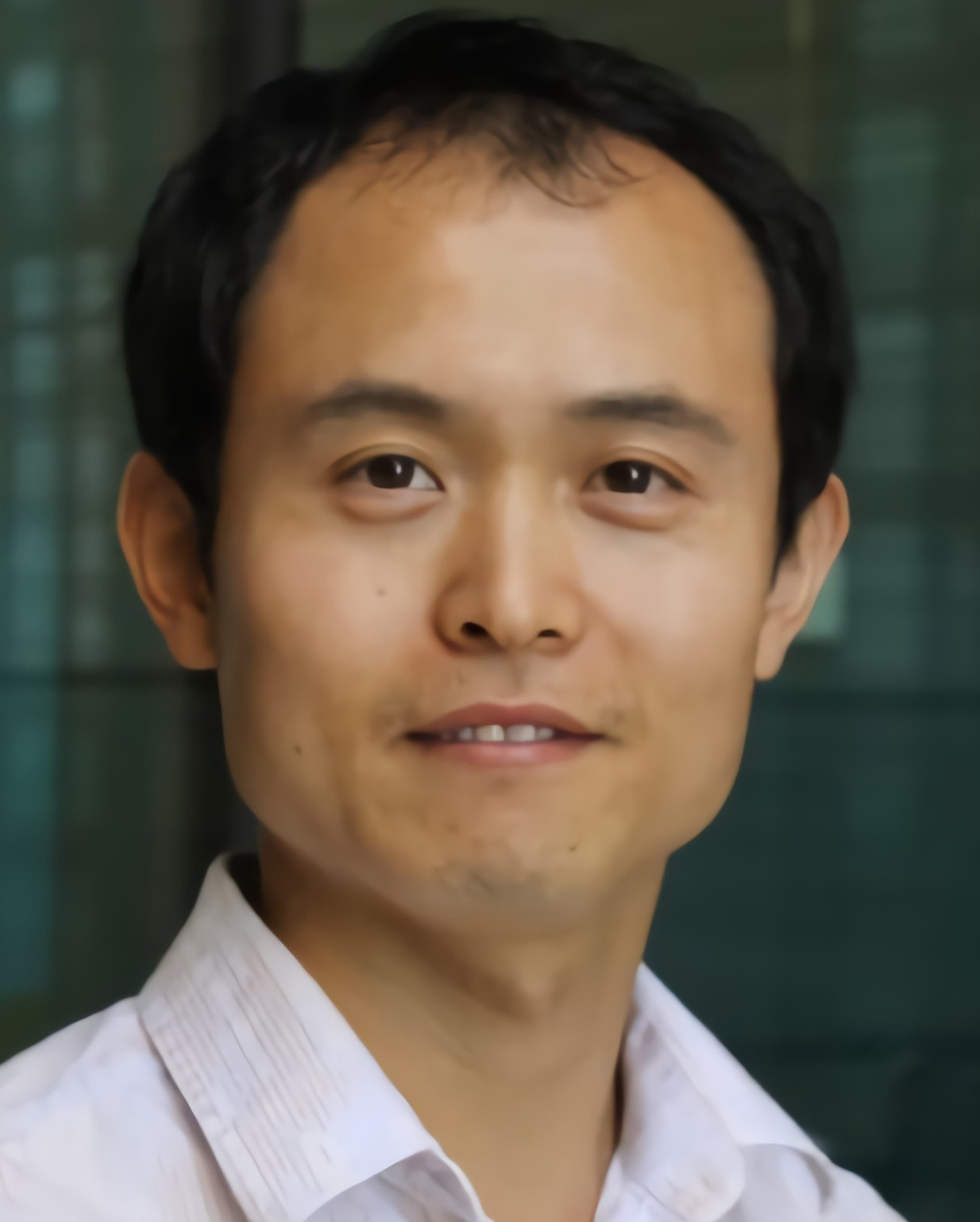}}]{Liangliang Nan} received the B.S. degree in material science and engineering from the Nanjing University of Aeronautics and Astronautics, Nanjing, China, in 2003, and the Ph.D. degree in mechatronics engineering from the Graduate University of the Chinese Academy of Sciences, Beijing, China, in 2009.

 From 2009 to 2013, he was an Assistant and then
 an Associate Researcher at the Shenzhen Institute of Advanced Technology (SIAT), Chinese Academy of Sciences, Beijing. From 2013 to 2018, he worked at the Visual Computing Center, King Abdullah University of Science and Technology (KAUST), Thuwal, Saudi Arabia, as a Research Scientist. He is currently an Assistant Professor with the Delft University of Technology (TU Delft), Delft, The Netherlands, where he is leading the AI Laboratory on 3D Urban Understanding (3DUU). His research interests include computer graphics, computer vision, 3D geoinformation, and machine learning.
\end{IEEEbiography}
\vspace{-10 mm}
%\newpage

\begin{IEEEbiography}[{\includegraphics[width=1in,height=1.25in,clip,keepaspectratio]{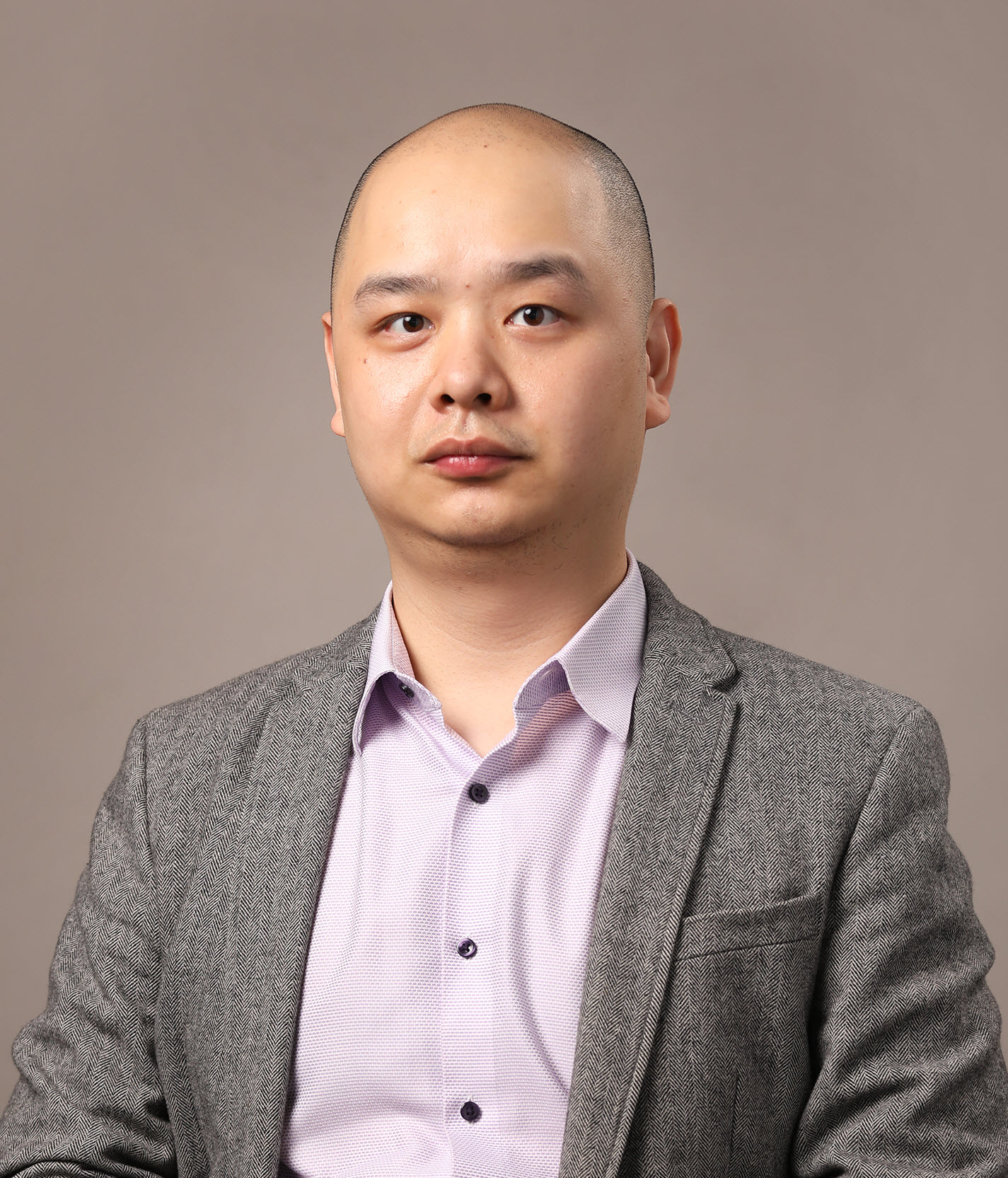}}]{Haoran Xie}
 (Senior Member, IEEE) received the
 Ph.D. degree in Computer Science from City
 University of Hong Kong, Hong Kong SAR, and Ed.D. degree in Digital Learning from University of Bristol, UK. He is
 currently an Associate Professor at the Department of
 Computing and Decision Sciences, Lingnan University,
 Hong Kong SAR. His research interests include
 artificial intelligence, big data, and educational technology.
 He has published 300 research publications,
 including 159 journal articles such as IEEE TPAMI,
 IEEE TKDE, IEEE TAFFC, IEEE TCVST, and so on.
He is the Editor-in-Chief of Natural Language Processing Journal, Computers \& Education: Artificial Intelligence and Computers \& Education: X Reality. He has been selected as The World Top 2\% Scientists by Stanford University.
 \end{IEEEbiography}
\vspace{-10 mm}
\begin{IEEEbiography}[{\includegraphics[width=1in,height=1.25in,clip,keepaspectratio]{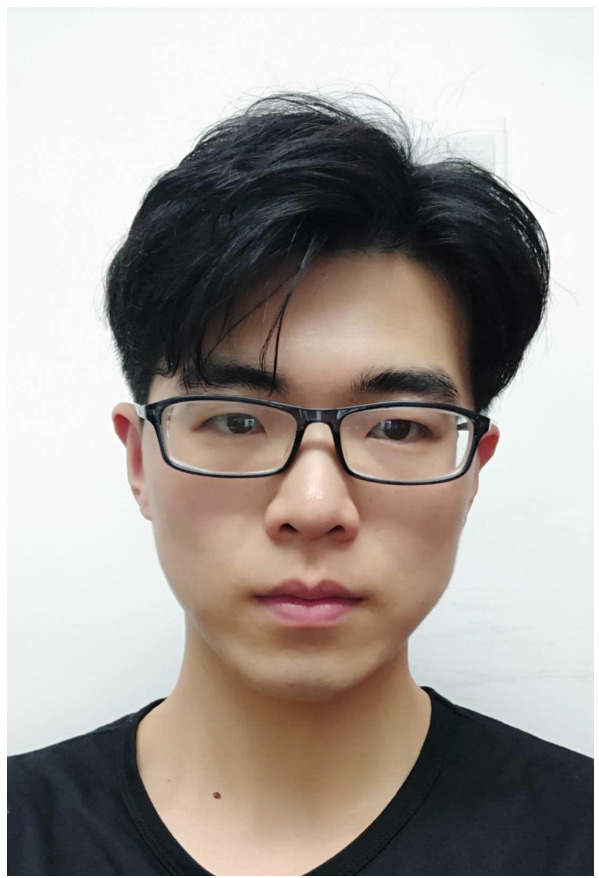}}]{Dening Lu}
received his BSc and MSc degrees in Electrical Engineering, both from the Nanjing University of Aeronautics and Astronautics (NUAA), China in 2018 and 2021, respectively. He is currently pursuing his Ph.D. degree in Systems Design Engineering with the Geospatial Sensing and Data Intelligence Group at the University of Waterloo, Canada. His research interests include 3D point cloud processing and deep learning. He has published papers in the IEEE Transactions on Instrumentation and Measurement and ICCV.
\end{IEEEbiography}
\vspace{-10 mm}
\begin{IEEEbiography}[{\includegraphics[width=1in,height=1.25in,clip,keepaspectratio]{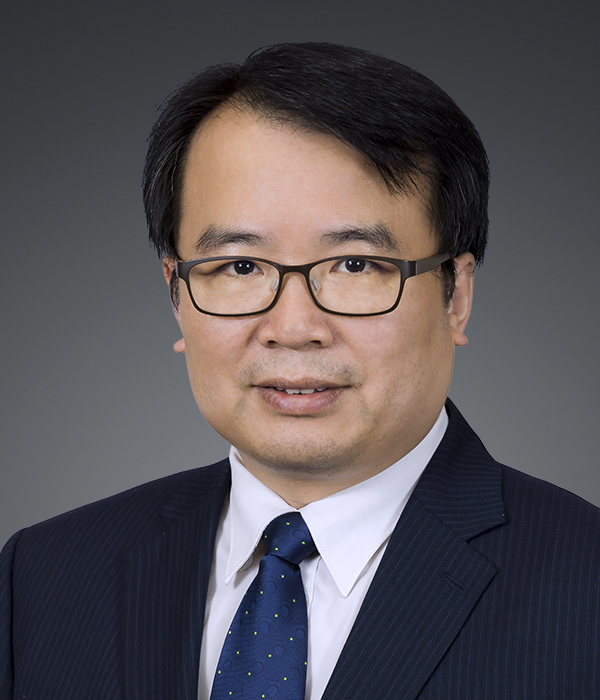}}]{Fu Lee Wang} (SM’15) received the B.Eng. degree in computer engineering and the M.Phil. degree in computer science and information systems from the University of Hong Kong, Hong Kong, and the Ph.D. degree in systems engineering and engineering management from the Chinese University of Hong Kong, Hong Kong. 
Prof. Wang is the Dean of the School of Science and Technology, Hong Kong Metropolitan University, Hong Kong. He has over 250 publications in international journals and conferences and led more than 20 competitive grants with a total greater than HK\$20 million. His current research interests include educational technology, information retrieval, computer graphics, and bioinformatics. 
Prof. Wang is a fellow of BCS and HKIE and a Senior Member of ACM. He was the Chair of the IEEE Hong Kong Section Computer Chapter and ACM Hong Kong Chapter.
\end{IEEEbiography}
\vspace{-10 mm}
\begin{IEEEbiography}[{\includegraphics[width=1in,height=1.25in,clip,keepaspectratio]{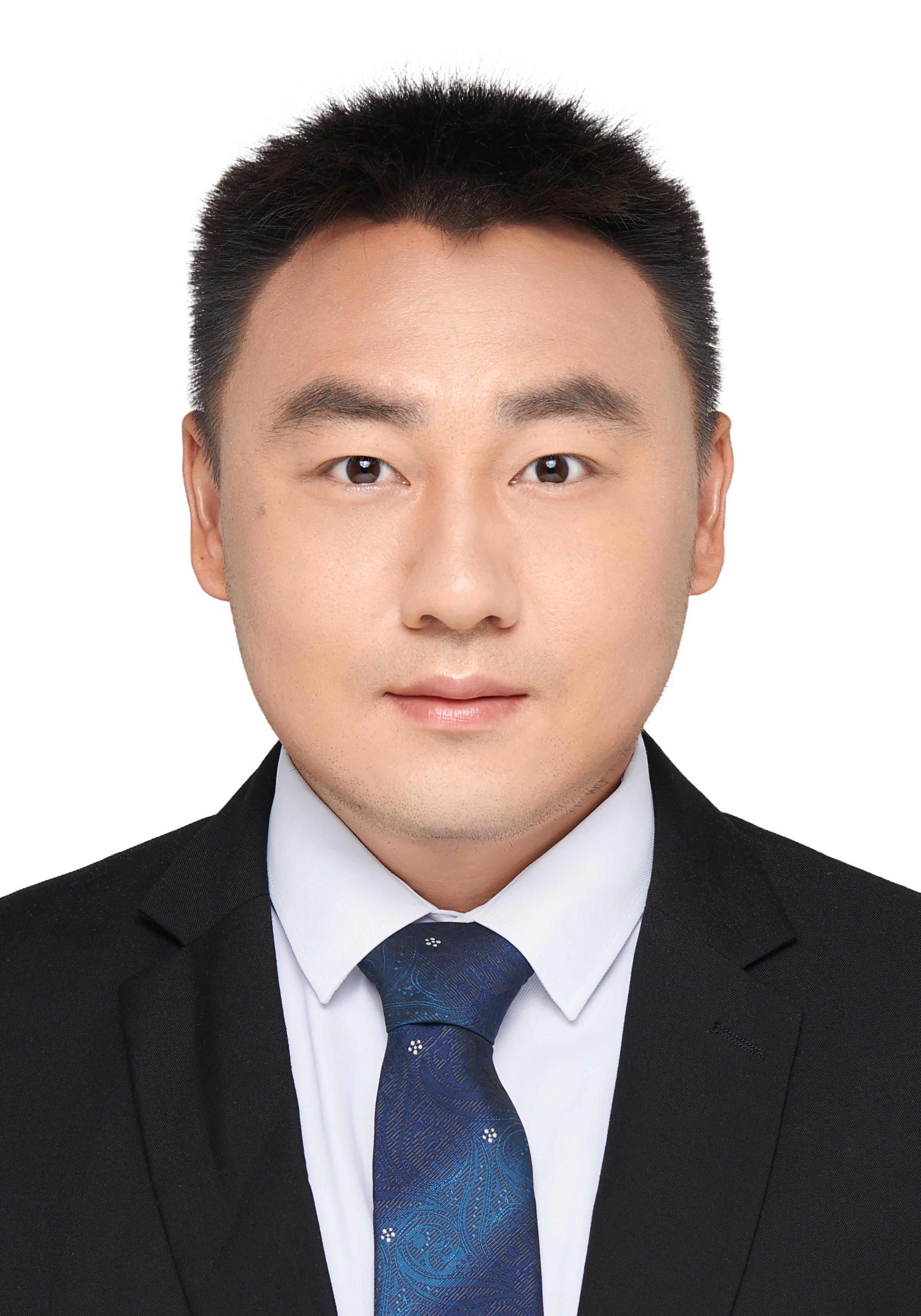}}]{Mingqiang Wei}
	received his Ph.D degree (2014) in Computer Science and Engineering from the Chinese University of Hong Kong (CUHK). He is a full Professor at the School of Computer Science and Technology, Nanjing University of Aeronautics and Astronautics (NUAA). Before joining NUAA, he served as an assistant professor at Hefei University of Technology, and a postdoctoral fellow at CUHK. He was a recipient of the CUHK Young Scholar Thesis Awards in 2014. He is now an Associate Editor for ACM TOMM, The Visual Computer (TVC), Journal of Electronic Imaging, and a leading Guest Editor for IEEE Transactions on Multimedia, and TVC. He has published
140 research publications, including TPAMI, SIGGRAPH, TVCG, CVPR, ICCV, et al. His research interests focus on 3D vision, computer graphics, and deep learning.
\end{IEEEbiography}


%\vfill


\end{document}